%% file: main.tex
\documentclass[10pt]{article} 
\usepackage[preprint]{tmlr}

\input{math_commands.tex}

\usepackage{hyperref}
\usepackage{url}
\usepackage{graphicx}
\usepackage{subfigure}
\usepackage{tabularx}
\usepackage{booktabs}
\usepackage{booktabs}
\usepackage{dcolumn}
\usepackage{makecell}
\usepackage{rotating} 
\usepackage{multirow}
\usepackage{paralist}
\usepackage{xcolor}
\usepackage{float}
\title{GANs Conditioning Methods: A Survey}

\author{\name Anis Bourou \email anis.bourou@ens.psl.eu \\
      \addr ENS, Université de Paris Cité\\
      \AND
      \name Valérie Mezger \email valerie.mezger@gmail.com\\
      \addr Université de Paris Cité
      \AND
      \name Auguste Genovesio \email auguste.genovesio@ens.psl.eu \\
      \addr ENS
      }

\def\month{MM}  
\def\year{YYYY} 
\def\openreview{\url{https://openreview.net/forum?id=XXXX}} 

\begin{document}

\maketitle

\begin{abstract}
In recent years, Generative Adversarial Networks (GANs) have seen significant advancements, leading to their widespread adoption across various fields. The original GAN architecture enables the generation of images without any specific control over the content, making it an unconditional generation process. However, many practical applications require precise control over the generated output, which has led to the development of conditional GANs (cGANs) that incorporate explicit conditioning to guide the generation process. cGANs  extend the original framework by incorporating additional information (conditions), enabling the generation of samples that adhere to that specific criteria. Various conditioning methods have been proposed, each differing in how  they integrate the conditioning information into both the generator and the discriminator networks.
In this work, we review the conditioning methods proposed for GANs, exploring the characteristics of each method and highlighting their unique mechanisms and theoretical foundations. Furthermore, we conduct a comparative analysis of these methods, evaluating their performance on various image datasets. Through these analyses, we aim to provide insights into the strengths and limitations of various conditioning techniques, guiding future research and application in generative modeling.

\end{abstract}

\section{Introduction}

Generative Adversarial Networks (GANs)~\cite{gan_goodfellow} are implicit generative model in which the data distribution is learned by comparing real samples with generated ones. This approach leverages an adversarial process between two neural networks: a generator, which produces synthetic data, and a discriminator, which evaluates the data to distinguish between real and generated (fake) samples. The competition between these networks drives both to improve, with the generator aiming to create increasingly realistic data while the discriminator becomes better at identifying fakes. Since their introduction, GANs have inspired numerous extensions and enhancements. Notably, the Deep Convolutional GAN (DCGAN)~\cite{dcgan}, marked a significant advancement by employing convolutional layers. DCGAN demonstrated the ability to generate high-quality images and contributed to the robustness of GAN training.
To address the inherent training difficulties and instability of GANs, different objectives were proposed~\cite{wgan, wgan_gp, object_1, object_2, object_3}, leading to more stable training and ultimately producing higher quality outputs. 
Self-Attention GANs (SAGAN)~\cite{selfattention} enhanced GANs' ability to capture global dependencies within images by integrating self-attention mechanisms~\cite{attention_paper}.
BigGAN~\cite{biggan}, scaled up the GAN architecture, achieving spectacular results on the ImageNet dataset through large batch sizes and careful architectural choices, pushing the boundaries of what GANs can achieve in terms of image quality and diversity. Furthermore,~\cite{progan} proposed the Progressive Growing GAN (ProGAN), a method to train GANs starting with low-resolution images and incrementally increasing the resolution as training progresses. This approach mitigated training instability and produced unprecedentedly high-resolution images. Additionally, StyleGAN~\cite{stylegans}, introduced a style-based generator that allows for fine-grained control over the generated image features, setting new benchmarks in image synthesis.
These foundational works have collectively expanded the application of GANs, including but not limited to image synthesis, data augmentation~\cite{gan_augmentation}, super-resolution~\cite{super_resolutions_gans}, and even creative domains such as art generation~\cite{gan_art_generation}.
Controlling the generative process of images is crucial for many applications such as image editing~\cite{Gauthier2015ConditionalGA,faceaging,invertible_cgan_image_editing,maskguided_conditional}, text-based image generation~\cite{text_to_image,text_to_imgae_hiera,text_tophoto_realistic}, 3D scene manipulation~\cite{3d_manipulation}, time series analysis~\cite{timeseriescond} and medical imaging~\cite{conditionalgan_medical_imaging1, conditionalgan_medical_imaging2,bourou2023}. 
Although cGANs were initially mentioned as a straightforward extension of GANs~\cite{gan_goodfellow}, they were first formally introduced in~\cite{mirza_conditional_gan}. Their approach introduced conditioning by incorporating a class embedding variable, which was concatenated with the input data fed into both the generator and the discriminator. The Auxiliary Classifier GAN (AC-GAN)~\cite{acgan_bias} 
advanced this concept by adding an auxiliary classifier to the discriminator, enabling it to distinguish between different classes of images. While AC-GAN improved the quality of conditional generation, it tended to produce images that were easy to classify~\cite{acgan_bias}. Moreover, as the number of classes increased, AC-GAN was prone to early collapse during the training process~\cite{acgan_bias, unbiased_acgan}.

To address these issues, several subsequent works introduced significant improvements. In~\cite{projectioncgans} proposed to perform the conditioning using a projection-based discriminator, achieving remarkable synthesis results, a principle later adopted by other cGAN architectures such as StyleGANs~\cite{stylegans,stylegan2,stylegan3} and BigGAN~\cite{stylegans,biggan}. ContraGAN~\cite{contragan} used contrastive loss to better capture data-to-data relations, improving the conditioning process.

Despite the numerous advancements and variations in conditional GANs, no prior surveys have comprehensively discussed the different architectures used in conditional GANs. This work aims to fill this gap by providing an extensive overview of the methods used to condition GANs, comparing their effectiveness across different datasets, and evaluating their robustness and performance.

The adversarial training scheme for GANs~\cite{gan_goodfellow} is described by the following loss function:
\begin{equation}
 \label{gan_good}
 \min_{G}\max_{D}\mathbb{E}_{x\sim p(x)}[\log{D(x)}] +  \mathbb{E}_{z\sim p_{\text{z}}(z)}[ \log{(1 - D(G(z)))}]
\end{equation}

Where $G$ and $D$ are, respectively, a parameterized generator and discriminator, $p(x)$ is the real data distribution, and $p_{z}(z)$ is a multivariate random distribution.  

GANs can be made conditional by considering additional information such as class labels, text, image, or other modalities. While conditioned, the GAN objective\ref{gan_good} can be reformulated as:

\begin{equation}
 \min_{G}\max_{D}\mathbb{E}_{x\sim p_{\text{}}(x)}[\log{D(x|y)}] +  \mathbb{E}_{z\sim p(z)}[\log{(1 - D(G(z|y)))}]
\end{equation}

In this survey, we focus on works that use class labels for conditioning, though many of the techniques discussed can be extended to other modalities. Our work is organized as follows: First, we outline the methods proposed for injecting class labels into the discriminator. Next, we describe the techniques suggested for conditioning the generator. Finally, we provide a comparison to evaluate and contrast the proposed conditional GANs.

\section{Discriminator conditioning approaches}\label{discriminator_condition}

The discriminator in GANs plays a crucial role, by providing feedback on the quality of the generated data samples to the generator. In the conditional setting, the discriminator should be provided with the class label, the earliest cGANs frameworks fed the class label $y$ to the discriminator by simply concatenating it with the feature vector~\cite{mirza_conditional_gan}. Variants of this method have proposed concatenating the class label embedding with the feature vector at different depths in the network \cite{conditional_concat_4,text_tophoto_realistic,temporal_gan_svc,dumoulin2017adversarially,invertible_cgan_image_editing,laplacegan_denton,reed_text_to_image}.

\begin{figure}[h]
    \centering
    \includegraphics[width=0.5\linewidth]{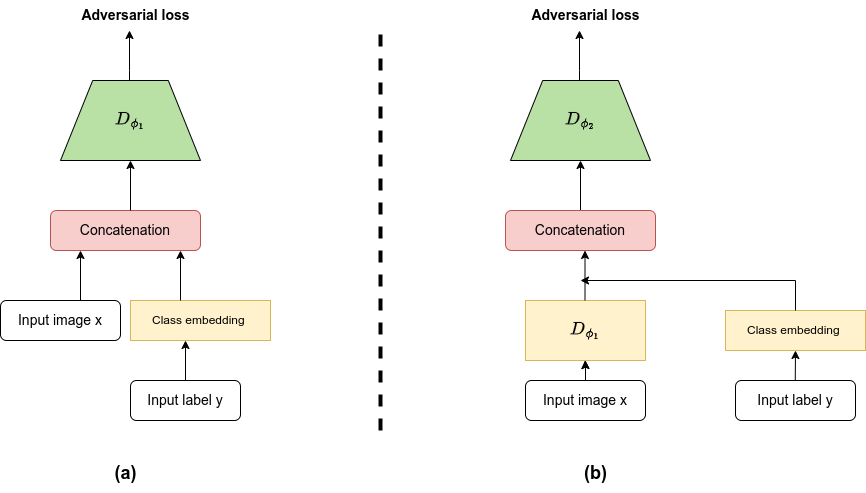}
    \caption{Conditioning by label embedding concatenation at different levels of the discriminator, (a) was proposed in~\cite{mirza_conditional_gan} and (b) in~\cite{reed_text_to_image} }
    \label{fig:subfigures}
\end{figure}

Subsequent works introduced new techniques for injecting the class label information into the discriminator. Depending on the conditioning method, we can group the cGAN discriminators into: \textbf{auxiliary classifier-based discriminators}, \textbf{projection-based discriminators} and \textbf{constrastive learning-based discriminators}. In the subsequent sections, we will delve into each method, outlining their respective mechanisms.

\subsection{Auxiliary-classifier based discriminators}

Concatenating class label information with the input image features can achieve conditioning; however, this approach is relatively simplistic and arbitrary, which may hinder GANs from accurately approximating the true data distribution. 
In this section, we present a collection of methods that condition the discriminator by incorporating an auxiliary classifier.

\subsubsection{Auxiliary classifier GAN (AC-GAN)}
The development of conditional Generative Adversarial Networks (cGANs) advanced significantly with the introduction of 
\textbf{Auxiliary Classifier GANs (AC-GANs)}~\cite{acgan}. This approach integrates an auxiliary classifier into the GAN discriminator to predict the class label of the generated image. This design shift, motivated by the potential for enhanced performance through multi-task learning, enables the AC-GAN to generate higher quality and class-specific images. Unlike prior cGANs, where conditioning information is directly fed to the discriminator via concatenation, AC-GAN employs a dual objective function. The first part, $L_S$ (Eq. \ref{ls}), focuses on the log-likelihood of correctly identifying real versus fake images, while the second part, $L_C$ (Eq. \ref{lc}), concentrates on accurately classifying these images into their respective classes:

\begin{equation}
\label{ls}
 L_{S}=\mathbb{E}_{x\sim p(x)}[\log{D(x)}] +  \mathbb{E}_{z\sim p_{\text{z}}(z), y\sim p_{\text{y}}(y) }[ \log{(1 - D(G(z,y)))}]
\end{equation}

\begin{equation}
\label{lc}
L_C = -\mathbb{E}_{x \sim p(x,y)}[\log C(x,y)] - \mathbb{E}_{z \sim p_z(z), y \sim p_y(y) }[\log(C(G(z,y)),y)]
\end{equation}

Where $C$ is the introduced auxiliary classifier, By combining $L_S$ and $L_C$ we obtain the AC-GAN loss:

\begin{equation}
\begin{aligned}
\min_{G,C} \max_D \mathcal{L}_{AC}(G,D,C) = & \mathbb{E}_{x \sim p_x(x)}[\log D(x)] + \mathbb{E}_{z \sim p_z(z), y \sim p_y(y) }[\log(1 - D(G(z,y)))] \\
& - \lambda_c \mathbb{E}_{ x \sim p(x,y)}[\log C(x,y)] - \lambda_c \mathbb{E}_{z \sim p_z(z), y \sim p_y(y) }[\log(C(G(z,y),y))]
\end{aligned}
\end{equation}
where $\lambda_C$ is a hyperparameter. 

Despite its advancements, \textbf{AC-GAN} suffers from a lack of diversity in the generated images, particularly as the number of classes increases. This issue arises from the model's tendency to generate images that are easier for the auxiliary classifier to categorize. This bias was explored in \cite{acgan_bias}, where \textbf{AC-GAN} was described as the Lagrangian of a constrained optimization problem that rejects the sampling of points near the classifier decision boundaries.

\subsubsection{Twin Auxiliary Classifier GAN (TAC-GAN)}

To address the limitations of \textbf{AC-GAN}, the \textbf{Twin Auxiliary Classifier GAN (TAC-GAN)} was proposed by~\cite{twin_auxiliary}, introducing an additional auxiliary classifier. In~\cite{twin_auxiliary}, it was shown that the absence of the negative conditional entropy term $-H_{q}(y|x)$ in the objective function of \textbf{AC-GAN} can lead to a degenerate solution that causes the generated images to be confined by the decision boundaries of the auxiliary classifier. This behavior explains the low intra-class diversity observed in the images synthesized by \textbf{AC-GAN}. To alleviate this issue, the authors proposed to add an additional classifier to the \textbf{AC-GAN} that predicts the class of the generated images. This additional auxiliary classifier $C^{mi}$ is trained to compete with the generator, optimizing the following objective function: 

\begin{equation}
\label{aux_classi2}
 \min_{G}\max_{C^{mi}}{V}(G,C^{mi})=\mathbb{E}_{z\sim p_{z}(z), y\sim p_{y}(y) }[\log{C^{mi}(G(z,y),y)}]
\end{equation}

combining Eq. \ref{aux_classi2} with the original AC-GAN objective leads to the total loss of TAC-GAN that reads: 

\begin{equation}
\label{aux_classi_twin_total}
\min_{G,C} \max_{D,C^{\text{mi}}} \mathcal{L}_{TAC}(G,D,C,C^{\text{mi}}) = \mathcal{L}_{AC}(G,D,C) + \lambda_{ac} V(G,C^{\text{mi}})
\end{equation}
It was shown in~\cite{twin_auxiliary}, theoretically and experimentally that this new loss function can be helpful in learning an unbiased distribution and generating more diverse images.

\subsubsection{Unbiased Auxiliary Classifier GAN (UAC-GAN)}

Similar to~\cite{twin_auxiliary}, \cite{unbiased_acgan} demonstrated that the lack of diversity observed in \textbf{AC-GAN} is induced by the absence of $-H_{q}(y|x)$ in the AC-GAN objective function. Furthermore, it was shown that the \textbf{TAC-GAN} can still converge to a degenerate solution. In addition to that, it was observed that using an additional classifier can lead to an unstable training~\cite{causalgan,unbiased_acgan}. Instead of using an additional classifier to minimize $-H_{q}(y|x)$, \cite{unbiased_acgan} proposed to estimate the mutual information $I_q(x;y)$ since:
\begin{equation}
\label{mutual_information}
I_q(x;y) = H_q(y)-H_q(y|x) = H_q(x)-H_q(x|y)
\end{equation}

To estimate $I_q(x;y)$, they employed the \textit{Mutual Information Neural Estimator(MINE)}~\cite{belghazi2021mine}. MINE is built on top of the Donsker and Varadhan bound~\cite{donsker}, $I_{Q}$ can be estimated using the following equation: 
\begin{equation}
 I_{q}^{MINE}(x,y)=\max_{T}{V_{MINE}(G,T)} 
\end{equation}
where: 
\begin{equation}
\label{aux_classi}
 V_{MINE}(G,T)=\mathbb{E}_{z\sim p(z),y\sim p_(y)}[T(G(z,y),y)] - \log \mathbb{E}_{z\sim p(z),y\sim q(y)} \exp{(T(G(z,y), y)}
\end{equation}
$T$ is a scalar-valued function that can be parameterized by a deep neural network. The final objective function is given by:

\begin{equation}
\label{aux_classi}
\min_{G,C} \max_{D,T}(G,D,C,T) = L_{AC}(G,D,C) + \lambda_{m} V_{MINE}(G,T)
\end{equation}

where $\lambda_{m}$ is a hyperparameter.

\cite{unbiased_acgan} demonstrated that directly estimating the mutual information effectively addresses the lack of diversity in \textbf{AC-GAN} without the need for an additional classifier, which can often lead to unstable training.

\subsubsection{Auxiliary Discriminative Classifier GAN (ADC-GAN)}

In another work,~\cite{discriminative_ac} proposed \textbf{Auxiliary Discriminative Classifier GAN (ADC-GAN)} to overcome the limitations of \textbf{AC-GAN}. They demonstrated that, for a fixed generator, the optimal classifier of \textbf{AC-GAN} is agnostic to the density of the generated distribution $q(x)$. Furthermore, they highlighted that the generators in \textbf{TAC-GAN} and \textbf{AC-GAN} optimize contradictory learning objectives as shown in Table~\ref{tab:divergences_table}.

\begin{table}[t]
    \label{tab:divergences_table}
    \centering
    \caption{learning objective for the generator under the optimal discriminator and classifier.}
    \vspace{10pt}
    \begin{tabular}{c|c}
        \hline
        \textbf{Method} & \textbf{Learning Objective for the Generator} \\
        \hline
        AC-GAN & $\min\limits_{G} JS(p_x\|q_x) + \lambda_1 KL(q_{x,y}\|p_{x,y})) - KL(q_x\|p_x)) + H_q(y|x))$ \\
        \hline
        TAC-GAN & $\min\limits_{G} JS(p_x\|q_x)) + \lambda_1 KL(q_{x,y}\|p_{x,y})) - KL(q_x\|p_x))$ \\
        \hline
        ADC-GAN & $\min\limits_{G} JS(p_x\|q_x) + \lambda_1 KL(q_{x,y}\|p_{x,y})$ \\
        \hline
    \end{tabular}
\end{table}

To alleviate these shortcomings, \textbf{ADC-GAN} uses a classifier that is able to classify the real data and the generated data separately. Using such a classifier $C_d:\textit{X} \rightarrow \textit{Y}^{+} \cup \textit{Y}^{-}$ ($\textit{Y}^{+}$ for real data and $\textit{Y}^{-}$ for generated data), the generator is encouraged to produce classifiable samples that look like the real ones. The objective functions for the discriminator, the discriminative classifier and the generator are:  

\begin{equation}
\label{aux_classi}
\begin{split}
 &\max_{D,C_d}{V_{AC}}(G,D) + \lambda(\mathbb{E}_{x,y\sim p_{x,y}}[\log C_{d}(y^+|x)] +\mathbb{E}_{x,y\sim q_{x,y}}[\log C_{d}(y^-|x)])\\
 &\min_{G}{V_{AC}}(G,D) - \lambda(\mathbb{E}_{x,y\sim p_{x,y}}[\log C_{d}(y^+|x)] -\mathbb{E}_{x,y\sim q_{x,y}}[\log C_{d}(y^-|x)])
\end{split}
\end{equation}

where:

$C_d(y^+|x) = \frac{\exp(\varphi^+(y) \cdot \phi(x))}{\sum_{\bar{y}} \exp(\varphi^+(\bar{y}) \cdot \phi(x)) + \sum_{\bar{y}} \exp(\varphi^-(\bar{y}) \cdot \phi(x))}$ and $C_d(y^-|x) = \frac{\exp(\varphi^-(y) \cdot \phi(x))}{\sum_{\bar{y}} \exp(\varphi^+(\bar{y}) \cdot \phi(x)) + \sum_{\bar{y}} \exp(\varphi^-(\bar{y}) \cdot \phi(x))}$

The function $\phi: \mathcal{X} \rightarrow \mathbb{R}^d$ serves as a feature extractor, transforming input data $X$ into a $d$-dimensional feature space. This feature extractor is shared with the original discriminator, which is represented as $D = \sigma \circ \psi \circ \phi$. Here, $\psi: \mathbb{R}^d \rightarrow \mathbb{R}$ is a linear mapping, and $\sigma: \mathbb{R} \rightarrow [0, 1]$ is a sigmoid function. Additionally, $\varphi^+: \mathcal{Y} \rightarrow \mathbb{R}^d$ and $\varphi^-: \mathcal{Y} \rightarrow \mathbb{R}^d$ are learnable embeddings capturing the label representations for real and generated data, respectively and $V_{AC}$ is the the original loss for \textbf{AC-GAN}.

\begin{figure}[h]
    \centering
    \includegraphics[width=0.9\linewidth]{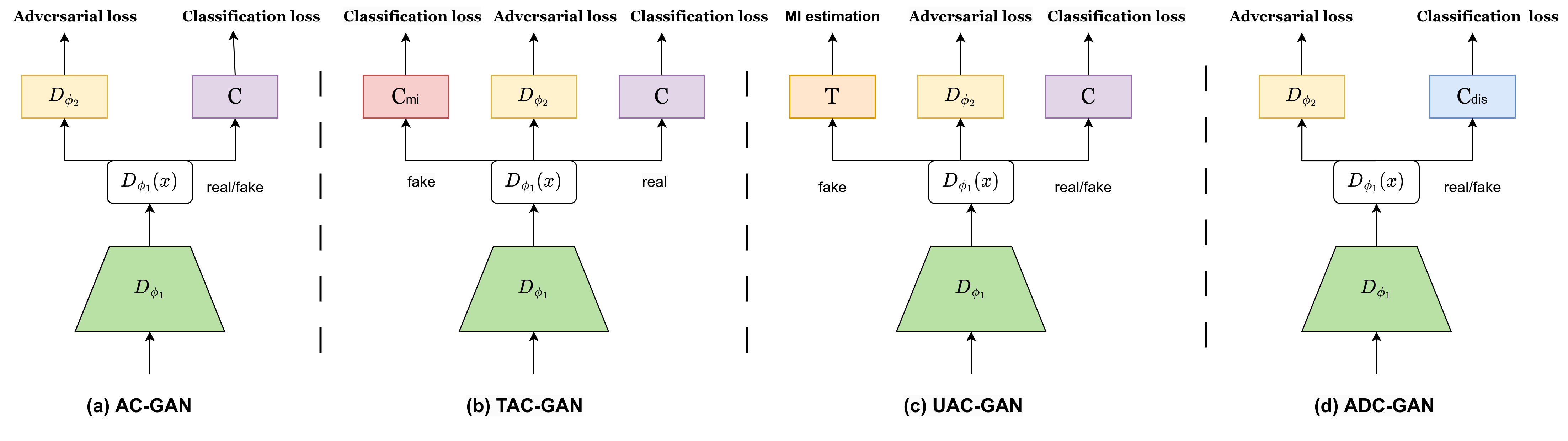}
    \caption{Auxiliary classifier based architectures: (a) AC-GAN, (b) TAC-GAN, (c) UAC-GAN, (d) ADC-GAN}
    \label{fig:subfigures}
\end{figure}

The authors of ADC-GAN proved that for a fixed generator, the optimal discriminative classifier is given as:  
$C_d^*(y^+|x) = \frac{p(x, y)}{p(x) + q(x)}, \quad C_d^*(y^-|x) = \frac{q(x, y)}{p(x) + q(x)}$ which shows that the optimal discriminative classifier is aware of the real and generated densities. Furthermore, they were able to discard the $-KL(p_x|q_x)$ term from the objective of the generator as shown in Table~\ref{tab:divergences_table}  In addition, they demonstrated that \textbf{ADC-GAN} leads to a more stable training and outperforms \textbf{AC-GAN} and \textbf{TAC-GAN} in terms of the quality of the generated images.

\subsection{Projection-based discriminators}\label{projection_discriminator_condition}

In the previous sections, conditioning the discriminator was achieved either by concatenating the class label or by adding an auxiliary classifier. Where the former can method can be very naive and sub-optimal in capturing the additional class label information, the latter can make the training more difficult and unstable. In~\cite{projectioncgans}, a new method for conditioning the discriminator was introduced by computing the inner product between the embedded conditional vector $y$ and the feature vector.

The design introduced in~\cite{projectioncgans} presents a novel method for cGANs by employing a projection discriminator, it was proposed by considering the optimal solution for the discriminator's loss function, \cite{projectioncgans} demonstrated that  under certain regularity assumptions, the discriminator's function can be reparameterized as follows:

\begin{equation}
f(x,y; \theta) =  f_1(x,y;\theta)+f_2(x;\theta) = y^TV \phi(x;\theta_\Phi) + \psi(\phi(x;\theta_\Phi); \theta_\Psi)
\end{equation}

where $V$ is the embedding matrix of $y$, $\phi(.,\theta_\Phi)$ is a vector output function of $x$, and $\psi(.,\theta_\Psi)$ is a scalar function. The learned parameters  $\theta = \{V, \theta_\phi, \theta_\psi\}$ are trained to optimize the adversarial loss.

The projection discriminator approach for conditional Generative Adversarial Networks (cGANs), as proposed in \cite{projectioncgans} offers notable improvements over traditional methods like concatenation. This technique enhances inter-class diversity, producing more varied and realistic samples across different classes, which is crucial in many applications. A significant advantage of this method is its avoidance of additional classifiers, leading to a greater training stability. The effectiveness and versatility of this approach are further evidenced by its adoption in various advanced GAN architectures, as seen in~\cite{biggan,crgan,icrgan,logan,sagan}.

BigGAN~\cite{biggan}, was among the pioneering GANs to employ discriminator projection techniques for conditional generation. It brought significant enhancements to the scaling of GAN training, enabling the generation of images with higher resolutions. A pivotal enhancement in BigGAN's design is the integration of orthogonal regularization, which contributed markedly to its improved performance. Furthermore, BigGAN drew inspiration from the Self-Attention GAN~\cite{selfattention}, particularly its utilization of self-attention blocks. These blocks aid both the discriminator and generator in more effectively capturing the global structure of images. Additionally, BigGAN's architecture facilitated the application of the truncation trick, which allows for nuanced balancing of the fidelity-diversity trade-off in generated images.

Another line of work that adopted the projection discriminator is the StyleGAN~\cite{stylegans,stylegan2,stylegan3}. The StyleGAN family of models represents a significant advancement in the use of projection discriminators. These models have achieved new state of the art results by incorporating innovative components like the Mapping Network and AdaIN normalization~\cite{adain}. Moreover, several techniques were introduced to enhance the quality of image generation, even when dealing with limited size data sets~\cite{adain,diffaug}.

\begin{figure}[h]
    \centering
    \includegraphics[width=0.2\linewidth]{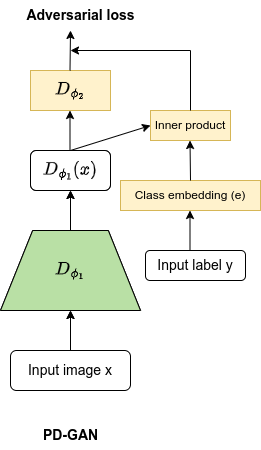}
    \caption{The architecture of the projection-based discriminator as proposed in~\cite{projectioncgans}.}
    \label{fig:projection_discriminator}
\end{figure}

\subsection{Contrastive learning based discriminators}
Contrastive learning~\cite{simclr,contrastive_1,contrastive1} mainly aims to develop deep, meaningful, and robust data representations. At its core, it involves training models to distinguish between pairs of examples that are either similar or dissimilar. During the training phase, the model is encouraged to draw closer the representations of similar items ('positive' pairs) while distancing those of dissimilar items ('negative' pairs). This approach not only strengthens the model's capacity to discern underlying data structures and patterns but also enhances generalization across various tasks. The effectiveness of contrastive learning is particularly evident in diverse domains such as computer vision \cite{contra_vision_1, contra_vision2,contra_vision3} and text processing \cite{contra_text1,contra_text2,contra_text3}. More recently, its application has been extended to generative models, as explored in~\cite{contragan,rebooting_ac_gan}, demonstrating its versatility and growing importance in the field of machine learning.

\subsubsection{Contrastive learning GAN (ContraGAN)}
ContraGAN~\cite{contragan} is a cGAN that achieve conditioning using a contrastive learning strategy by capturing the data-to-data relations. Indeed, it was suggested in~\cite{contragan} that the conditioning in \textbf{AC-GAN} and \textbf{ProjGAN} can only capture data-to-class relations of training examples while neglecting the data-to-data relations. To alleviate this,~\cite{contragan} have proposed the \textbf{Conditional Contrastive (2C) loss}, a self-supervised learning objective that controls the distances between embedded images depending on their respective labels.

The 2C loss can be seen as an adaptation of the NT-Xent loss~\cite{simclr}. Given a minibatch of training images $\textbf{X} = \{ \textbf{x}_1, \textbf{\ldots}, \textbf{x}_m\}, \text{ where } x \in \mathbb{R}^{W \times H \times 3}$ and their corresponding labels $\textbf{y}=\{y_1,\ldots,y_m\}$, an encoder $S(x) \in \mathbb{R}^{k}$, a projection layer $h : \mathbb{R}^k \rightarrow \mathbb{S}^d$  that embeds onto a unit hypersphere, the NT-Xent loss conducts random data augmentations $T$ on the training data $X$, denoted as $A =\{ \textbf{x}_1,T(\textbf{x}_1), \ldots,\textbf{x}_m,T(\textbf{x}_m)\} = \{\textbf{a}_1,\ldots,\textbf{a}_{2m}\}$, the loss is given by:

\begin{equation}
\label{nt_xent}
\ell(a_i, a_j; t) = -\log \left( \frac{\exp\left( \ell(a_i)^T \ell(a_j) / t \right)}{\sum_{k=1}^{2m} 1_{k \neq i} \exp\left( \ell(a_i)^T \ell(a_k) / t \right)} \right)
\end{equation}
where, $t$ is the temperature that controls the attraction and repulsion forces.

In~\cite{contragan} the discriminator network before the fully connected layer ($D_{\phi 1}$) is considered as the encoder network, an additional multi-layer perceptrons $h$ is used as a projection layer. Instead of using data augmentation, the authors used the embeddings of the class labels to capture the data-to-class relations, the modified loss is given as follows:

\begin{equation}
    \label{nt_xent_mod2}
\ell(x_i, y_i; t) = -\log \left( \frac{\exp(l(x_i)^Te(y_i)/t)}{\exp(l(x_i)^Te(y_i)/t) + \sum_{k=1}^{m} \mathbf{1}_{k \neq i} . \exp(l(x_i)^T l(x_k)/t)} \right)
\end{equation}

In order to ensure that the negative samples having the same label as $y_i$ are not apart, a cosine similarity of such samples is added to the numerator of Eq.~\ref{nt_xent_mod2} giving rise to the 2C loss: 

\begin{equation}
\label{2C}
l_{2C}(x_i, y_i; t) = -\log \left( 
\frac{\exp(l(x_i)^T e(y_i) / t) + \sum_{k=1}^{m}  \mathbf{1}_{y_k = y_i} \cdot \exp(l(x_i)^T l(x_k) / t)}
{\exp(l(x_i)^T e(y_i) / t) + \sum_{k=1}^{m}  \mathbf{1}_{k \neq i} \cdot \exp(l(x_i)^T l(x_k) / t)}
\right)
\end{equation}
where $l(x_i)$ is the embedding of the image $x_i$ and $e(y_i)$ the embedding of the class label $y_i$.
 
Eq.~\ref{2C} ensures that a reference sample \( x_i \) is drawn closer to its corresponding class embedding \( e(y_i) \) while distancing it from other classes.

By minimizing this 2C loss, \textbf{ContraGAN} effectively reduces the distance between embeddings of images with the same labels while increasing the distance between embeddings of images with different labels. This dual consideration of data-to-data \( l(x_i)^T l(x_k) \) and data-to-class \( l(x_i)^T e(y_i) \) relations marks a significant advancement over traditional methods.

\subsubsection{Rebooting Auxiliary Classifier GAN (ReACGAN)}
The introduction of contrastive learning in conditioning GANs paved the way for addressing data-to-data relations, a crucial aspect previously overlooked in previous work, particularly in classifier-based GANs like \textbf{ACGAN}. Building on this foundation, 
Rebooting ACGAN (ReACGAN)~\cite{rebooting_ac_gan} introduces the \textbf{Data-to-Data Cross-Entropy loss (D2D-CE)}. This novel approach specifically targets the early training collapse and the generation quality issues inherent in ACGAN. \cite{rebooting_ac_gan} started by considering the empirical cross-entropy loss used in ACGAN, which is given as follows: 

\begin{equation}
\mathcal{L}_{CE} = -\frac{1}{N} \sum_{i=1}^{N} \log \left( \frac{\exp(F(x_i)^T w_{y_i})}{\sum_{j=1}^{c} \exp(F(x_i)^T w_j)} \right)  
\end{equation}

where $F:\mathcal{X}\rightarrow\mathbb{R}^d$ $\text{is feature extractor and a single fully connected layer classifier } C:F\rightarrow \mathbb{R}^c$ which is parameterized
by $\textbf{W}=[w_1 \cdots w_c] \in \mathbb{R}^{d \times c}$, where $c$ is the number of classes. \cite{rebooting_ac_gan} found that at the early training stage the average norm of ACGAN's input features maps increases. Respectively, the average norm of the gradients dramatically increases at the early training steps and decreases with the high class probabilities of the classifier. In addition, it was observed that as the average norm of the gradients decreases, the FID value of ACGAN does not decrease indicating the collapse of ACGAN.

\cite{rebooting_ac_gan} found that normalizing the feature embeddings onto a unit hypersphere effectively solves the ACGAN’s early-training collapse. Specifically, the authors of ReACGAN introduced a projection layer $P$ on the feature extractor $F$ and they normalized both the feature embeddings $\frac{P(F(x_i))}{\|P(F(x_i))\|}$ (denoted as $\textbf{f}_i$ and the weight vector $\frac{w_{y_i}}{\|w_{y_i}\|}$  (denoted as $v_{y_i}$)

In addition to the normalization, \cite{rebooting_ac_gan} introduced  a contrastive loss \textbf{Data-to-Data Cross-Entropy(D2D-CE)} to better capture the data-to-data relations as in ContraGAN, furthermore they introduced two  margin values to the D2D-CE to guarantee inter-class separability and intra-class variations. The contrastive \textbf{D2D-CE} loss reads: 

\begin{equation}
\label{rebootacgan}
L_{D2D-CE} = -\frac{1}{N} \sum_{i=1}^{N} \log \left( \frac{\exp([f_i^T v_{y_i} - m_p]_-/\tau)}{\exp([f_i^T v_{y_i} - m_p]_-/\tau) + \sum_{j \in N(i)} \exp([f_i^T f_j - m_n]_+/\tau)} \right)
\end{equation}

where, \(\tau\) is the temperature parameter, and \(N(i)\) denotes the set of indices for negative samples with labels different from the reference label \(v_{y_i}\) in a batch. Margins \(m_p\) and \(m_n\) are used to manage similarity values for easy positives and negatives, respectively. The terms \([.]_{-}\) and \([.]_{+}\) correspond to the \(\min(.,0)\) and \(\max(.,0)\) functions.

This contrastive loss function proved to be effective in overcoming the limitations of ACGAN, significantly enhancing both class consistency and image diversity.

\begin{figure}[h]
    \centering
    \includegraphics[width=0.7\linewidth]{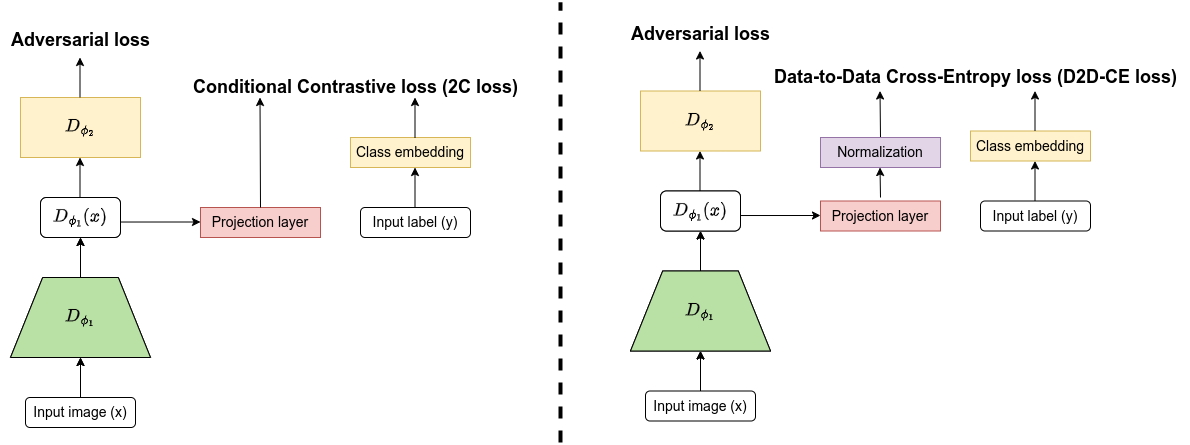}
    \caption{Contrastive learning based architectures. On the left: the discriminator architecture of ContraGAN. On the right: the Reboot AC-GAN architecture.}
    \label{fig:subfigures}
\end{figure}

In contrast to the \textbf{2C} loss, the \textbf{D2D-CE} objective does not hold false positives in the denominator, which can cause unexpected repulsion forces. Furthermore, introducing the margins in \textbf{D2D-CE} loss 
can prevent having large gradient that can be caused by pulling easy positive samples.

\subsection{Towards a unified framework for conditioning the discriminator}

As previously discussed, various methods have been introduced to condition the discriminator, either by incorporating auxiliary classifiers or by employing alternative approaches. While the inclusion of a classifier in \textbf{ACGAN} effectively achieved conditioning, alternative approaches have successfully conditioned the discriminator without the need for a classifier. In~\cite{unified_cgan}, showed that the use of classifiers can benefit conditional generation. Furthermore, they introduced a unifying framework named \textbf{Energy-based Conditional Generative Adversarial Networks (ECGAN)} which explains several cGAN variants. In order to connect the classifier-based and the classifier-free approaches used equivalent formulations of the joint probability $p(x,y)$. 

From a probabilistic perspective, a classifier can be seen as a function that approximates $p(y|x)$, the probability that $x$ belongs to $y$. Similarly, a conditional discriminator can be viewed as a function that approximates $p(x|y)$, the probability that $x$ is real given a class $y$, the joint probability is given as follows: 
\begin{equation}
\label{joint}
    \log p(x,y) = \log p(x|y) + \log p(y)
                = \log (y|x)  + \log p(x)
\end{equation}

In Eq.\ref{joint} we observe that the joint probability distribution $\log p(x,y)$ can be approached through two distinct methods. The first method involves modeling a conditional discriminator $p(x|y)$, while the second focuses on a classifier $p(y|x)$. By sharing the parameterization between these models, the training process becomes mutually beneficial, allowing improvements in the conditional discriminator to enhance the classifier's performance, and vice versa.

\subsubsection{Approaching the joint distribution from the conditional discriminator perspective}

Similar to the energy based models~\cite{energy_based},  $log$ $p(x,y)$ was parameterized in~\cite{unified_cgans} using a function $f_\theta(x)$, where $\exp(f_\theta(x)[y]) \propto p(x, y)$.

Therefore, the log-likelihood can be modeled as follows:

\begin{equation}
\label{jd_cond}
\log p_{\theta}(x|y) = \log \left( \frac{\exp \left( f_{\theta}(x)[y] \right)}{Z_y(\theta)} \right) = f_{\theta}(x)[y] - \log Z_y(\theta),
\end{equation}

Where $Z_y(\theta) = \int_{x'} \exp \left( f_{\theta}(x')[y] \right) dx'$

Optimizing Eq. \ref{jd_cond} is intractable because of the partition function $Z_y(\theta)$. By introducing the Fenchel duality 
of the partition function $Z_y(\theta)$ and a trainable generator $q_\phi(z,y)$, where $z \sim \mathcal{N}(0,1)$,~\cite{unified_cgan} showed that the maximum likelihood estimation in Eq.\ref{jd_cond} is:

\begin{table}[t]
\caption{Some reprsentative cGAN discriminator's architectures as an ECGAN approximation}
\label{table-approximations}
\begin{center}
\begin{tabular}{llll}
\multicolumn{1}{c}{\bf Existing cGAN}  &\multicolumn{1}{c}{\bf $\alpha$} &\multicolumn{1}{c}{\bf $\lambda_{clf}$} &\multicolumn{1}{c}{\bf $\lambda_{c}$}
\\ \hline \\
ProjGAN         &\ 0 &\ 0 &\ 0\\
AC-GAN       &\ 0 &\hspace{-0.7em}> 0 &\ 0 \\
ContraGAN      &\ 0 &\ 0 &\hspace{-0.7em}> 0 \\
\end{tabular}
\end{center}
\end{table}

\begin{equation}
\label{joint_1}
\max_{\theta} \min_{\phi} \sum_{y} \mathbb{E}_{p_d(x|y)} \left[ f_\theta(x)[y] \right] - \mathbb{E}_{p(z)} \left[ f_\theta(q_\phi(z, y))[y] \right] - H(q_\phi(\cdot, y))
\end{equation}

By discarding the entropy term, this equation has the form of the GAN.

The discriminator loss in this case is given by:
\begin{equation}
\mathcal{L}_{d_1}(x, z, y; \theta) = -f_\theta(x)[y] + f_\theta(q_\phi(z))[y]
\end{equation}

\subsubsection{Approaching the joint distribution from the classifier perspective}

As depicted in Eq.\ref{joint}, $log$ $p(x,y)$ can also be approximated using $log$  $p(y|x)$ and $log$ $p(x)$. Using the energy function introduced earlier $f_\theta(x)$, $log(p_\theta(y|x)$ can be expressed as:
\begin{equation}
p_\theta(y|x) = \frac{p_\theta(x, y)}{p_\theta(x)} = \frac{\exp(f_\theta(x)[y])}{\sum_{y'} \exp(f_\theta(x)[y'])},
\end{equation}

Which can be written using a softmax as:
\begin{equation}
\mathcal{L}_{\text{clf}}(x, y; \theta) = -\log \left( \text{SOFTMAX}\left(f_\theta(x)\right)[y] \right)
\end{equation}

Furthermore,~\cite{unified_cgan} showed that by setting $h_\theta(x) = \log \sum_y \exp(f_\theta(x)$, maximizing the log-likelihood of $p(x)$ is equivalent to solving the following optimization problem.

\begin{equation}
\label{joint_2}
\max_{\theta} \min_{\phi} \, \mathbb{E}_{p_d(x, y)} \left[ h_\theta(x) \right] - \mathbb{E}_{p(z)} \left[ h_\theta(q_\phi(z, y)) \right] - H(q_\phi)
\end{equation}

Similar to Eq. ~\ref{joint_1}, we can see that the Eq.~\ref{joint_2} has the form of the traditional GAN. In this case, the equation of the  discriminator is given as:

\begin{equation}
\mathcal{L}_{d_2}(x, z, y; \theta) = -h_\theta(x) + h_\theta(q_\phi(z))
\end{equation}

In~\cite{unified_cgan}, two approaches were proposed to estimate the entropy terms in Eq. \ref{joint_1} and Eq. \ref{joint_2}. The first approach involves considering the entropy term to be zero, based on the fact that entropy is always non-negative, the constant zero is a lower bound. The second approach involves estimating a variational lower bound. The authors demonstrated that the \textbf{2C} loss, introduced in ContraGAN, serves as a lower bound in this context.

\subsubsection{Unifying cGANs discriminators }

To unify the classifier-based and classifier-free discriminators, \cite{unified_cgan} proposed the ECGAN discriminator with the following objective:

\begin{equation}
L_D(x, z, y; \theta) = L_{d1}(x, z, y; \theta) + \alpha L_{d2}(x, z, y; \theta) + \lambda_c L_{\text{C}}^{real} + \lambda_{\text{clf}} L_{\text{clf}}(x, y; \theta)
\end{equation}

where:

\( L_{d1}(x, z, y; \theta) \) is designed for the conditional discriminator, adjusting the output specifically for class-corresponding data pairs \( (x, y) \). Conversely, \( L_{d2}(x, z, y; \theta) \) addresses the unconditional aspect, updating outputs based on the realness of \( x \), independent of the class. The classifier training component, \( L_{\text{clf}}(x, y; \theta) \), increases the output for the correct class \( y \) and decreases it for other classes, thus enhancing classification accuracy. Lastly, the component \( L_{\text{C}}^{real} \) which is the contrastive loss calculated on real samples, plays a crucial role in refining latent embeddings, by pulling closer the embeddings of data with the same class. The detailed derivation of this loss can be found in~\cite{unified_cgans}.

A significant aspect of the \textbf{ECGAN} framework presented in this paper is its ability to unify various representative variants of cGAN, including ACGAN, ProjGAN and ContraGAN. These variants are different approximations within the unified \textbf{ECGAN} framework, as shown in Table \ref{table-approximations}, demonstrating the versatility of \textbf{ECGAN}.

\section{Generator conditioning approaches}\label{generator_condition}

To condition the generator, most approaches typically involve directly integrating the label with the generator, either through concatenation or by employing some normalization techniques such as \textbf{conditional batch normalization} and \textbf{adaptive instance normalization}~\cite{acgan,twin_auxiliary,contragan,rebooting_ac_gan,discriminative_ac}. These methods effectively incorporate label information, enabling the generator to produce outputs that accurately reflect the desired attributes.

In this section, we explore the various techniques used for generator conditioning. The majority of popular methods rely on batch normalization, as the basis for conditioning. Initially, batch normalization~\cite{batch_normalization} was proposed to make the training of deep learning models faster and more stable. Given an input batch $x \in \mathbb{R}^{N \times C \times H \times W}$, each batch normalization layer has two learnable parameters, $\gamma_{batch}$ and $\beta_{batch}$ which shift and scale the normalized input, respectively:

\begin{equation}
z = \gamma_{batch} \left( \frac{x - \mu_{c}}{\sigma_{c}} \right) + \beta_{batch}
\end{equation}

where $\mu_c(x) = \frac{1}{NHW} \sum_{n=1}^{N} \sum_{h=1}^{H} \sum_{w=1}^{W} x_{nchw}$ and $\sigma_c(x) = \sqrt{\frac{1}{NHW} \sum_{n=1}^{N} \sum_{h=1}^{H} \sum_{w=1}^{W} (x_{nchw} - \mu_c(x))^2}$.

\textbf{Instance normalization} , proposed in~\cite{Ulyanov} as an alternative to batch normalization, was motivated by style transfer applications. In instance normalization, the standard deviation ($\sigma_{\text{instance}}(x)$) and mean ($\mu_{\text{instance}}(x)$) are computed for each individual instance, which can be considered as contrast normalization, whereas in batch normalization (BN), these statistics are computed across the entire batch. In~\cite{Ulyanov}, it was observed that significant improvements could be achieved using instance normalization.

In~\cite{dumoulin_artisitic_style} \textbf{conditional instance normalization} was introduced to learn different artistic styles with a single conditional network  where it takes a content image and a given style to apply and produces a pastiche corresponding to that style. The authors found that to model a style it is sufficient to specialize scaling and shifting parameters after normalization to each specific style,

Similarly, \textbf{conditional batch normalization } was used to condition vision systems on text. For instance, it was used in~\cite{batch_normalization_language} as an efficient technique to modulate convolutional feature maps by text embeddings.

In~\cite{adain}, \textbf{\textbf{Adaptive Instance Normalization (AdaIN)}} was introduced. AdaIN can be seen as an extension of instance normalization, where  the shift and the scale are not learnt but computed. Given an input content image $x$ and an input style image $y$ the affine parameters are computed as follows:

\begin{equation}
\text{AdaIN}(x, y) = \sigma_{instance}(y) \left( \frac{x - \mu_{instance}(x)}{\sigma_{instance}(x)} \right) + \mu_{instance}(y)
\end{equation}

By computing the the affine transformation, AdaIN aligns the channel-wise mean and variance of the input image $x$ to match those of the style image $y$. the authors showed that $AdaIN$ lead to better style transfer compared to the other methods. Furthermore, it was extensively used in the StyleGAN family of models~\cite{stylegans,stylegan2,stylegan3}

In~\cite{film}, the authors introduced a general purpose method for conditioning a neural network on text embeddings called Feature-wise Linear Modulation (FILM). FILM learns functions f and h which output $\gamma_i,c$ and $\beta_i,c$ to modulate a neural network's activation $F_i,c$, the feature-wise affine transformation is given by:

\begin{table}
  \label{tab:cifar10}
  \centering
  \small 
  \begin{tabularx}{\textwidth}{@{}lXXXXXX@{}} 
    \toprule
    \textbf{Method} & \textbf{FID $\downarrow$} & \textbf{IS $\uparrow$} & \textbf{Coverage $\uparrow$} & \textbf{Density $\uparrow$} & \textbf{Recall $\uparrow$} & \textbf{Precision $\uparrow$}\\
    \midrule
    AC-GAN & 33.31 $\pm 1.8$ & 6.82 $\pm 0.79$ & 0.39 $\pm 0.04$ & 0.57 $\pm 0.02$ & 0.21$\pm 0.06$ & 0.63 $\pm 0.01$ \\
    ProjeGAN & 32.14 $\pm 1.87$ & 7.08 $\pm 0.42$ & 0.39 $\pm 0.021$ & 0.57 $\pm 0.032$ & 0.26 $\pm 0.04$ & 0.63 $\pm 0.02$ \\
    TAC-GAN & 181 $\pm 12.88$ & 5.46 $\pm 0.85$ & 0.02 $\pm 0.01$ & 0.084 $\pm 0.02$ & 0.07 $\pm 0.06$ & 0.25 $\pm 0$ \\
    BigGAN & 5.44 $\pm 0.12$ & 9.63 $\pm 0.08$ & \textbf{0.87} $\pm 0.003$ & 0.99 $\pm 0.01$ &  0.62$\pm 0.002$ & 0.74 $\pm 0.001$ \\
    ADC-GAN & 5.06 $\pm 0.19$  & \textbf{9.95} $\pm 0.06$ & 0.86 $\pm 0.01$  & 0.89$\pm 0.01$  & 0.66 $\pm 0.01$  & 0.71 $\pm 0.01$  \\
    ContraGAN &  5.99$\pm 0.91$ & 9.5 $\pm 0.21$ & 0.84 $\pm 0.011$ & 0.95 $\pm 0.01$ & 0.60 $\pm 0.004$ & 0.74 $\pm 0.002$ \\
    Reboot AC-GAN & 5.65 $\pm 0.18$ & 9.61 $\pm 0.09$ & 0.85 $\pm 0.004$ & \textbf{0.97} $\pm 0.01$ & 0.59 $\pm 0.002$ & \textbf{0.75} $\pm 0.01$ \\
    StyleGAN2 & \textbf{4.87$\pm 0.18$} & 8.1 $\pm 0.02$ & 0.86 $\pm 0.01$ & 0.82 $\pm 0.01$ & \textbf{0.68} $\pm 0.01$ & 0.7 \\
    \bottomrule
  \end{tabularx}
  \caption{Best FID score achieved by cGANs architectures during 80000 training steps on the CIFAR 10 dataset}
  \label{tab:cifar_10}
\end{table}

\begin{table}
  \label{tab:fid_carnivors}
  \centering
  \small 
  \begin{tabularx}{\textwidth}{@{}lXXXXXX@{}} 
    \toprule
    \textbf{Method} & \textbf{FID $\downarrow$} & \textbf{IS $\uparrow$} & \textbf{Coverage $\uparrow$} & \textbf{Density $\uparrow$} & \textbf{Recall $\uparrow$} & \textbf{Precision $\uparrow$}\\
    \midrule
    AC-GAN & 117.52  $\pm 4.58$ & 9.29 $\pm 0.80$ & 0.04 $\pm 0.01$ & 0.15 $\pm 0.02$ &  0.034$\pm 0.05$ & 0.15 $\pm 0.034$ \\
    ProjGAN & 181 $\pm 12.88$ & 5.46 $\pm 0.85$ & 0.02 $\pm 0.01$ & 0.084 $\pm 0.02$ & 0.07 $\pm 0.06$ & 0.25 $\pm 0$ \\
    TAC-GAN & 150.66 $\pm 9.9$ & 5.96 $\pm 0.84$ & 0.107 $\pm 0.14$ & 0.058 $\pm 0.02$ & 0.121 $\pm 0.04$ & 0.183 $\pm 0.05$\\
    BigGAN & 44.3 $\pm 7.54$ & 11.6 $\pm 0.80$ & 0.36 $\pm 0.07$ & 0.46 $\pm 0.06$ & 0.46 $\pm 0.05$ & 0.58 $\pm 0.05$ \\
    ADC-GAN & 15.11 $\pm 1.39$ & 15.46 $\pm 0.26$ & \textbf{0.715} $\pm 0.01$ & 0.76 $\pm 0.11$  & 0.51 $\pm 0.02$ & 0.69 $\pm 0.01$ \\
    ContraGAN & 22.95 $\pm 3.27$ & 12.89 $\pm 0.44$ & 0.49 $\pm 0.04$ & 0.75 $\pm 0.03$ & 0.32 $\pm 0.03$ & 0.73 $\pm 0.02$ \\
    Reboot AC-GAN & \textbf{12.55} $\pm 1.3$ & \textbf{15.93} $\pm 0.1$ & 0.696 $\pm 0.04$ & \textbf{0.99} $\pm 0.05$ & 0.33 $\pm 0.02$ & \textbf{0.8} $\pm 0.02$ \\
    StyleGAN2 & 16.99 $\pm 1.46$ & 14.85 $\pm 0.3$ & 0.67 $\pm 0.01$ & 0.67 $\pm 0.0173$ & \textbf{0.52} $\pm 0.01$ & 0.68 \\
    \bottomrule
  \end{tabularx}
  \caption{Best FID score achieved by cGANs architectures during 80000 training steps on the Carnivores dataset}
  \label{tab:cifar_100}
\end{table}

\begin{equation}
\text{FiLM}(F_{i,c}) = \gamma_{i,c} F_{i,c} + \beta_{i,c}
\end{equation}

FiLM is computationally efficient as it only requires two parameters per modulated feature map and it does not scale with the image resolution.

In BigGAN, a strategy similar to FiLM \cite{film} was used.
\section{Experiments}

\begin{figure}[t]
    \centering
    \begin{minipage}{0.48\textwidth}
        \label{fid_number_of_classes_afhq}
        \centering
        \includegraphics[width=\linewidth]{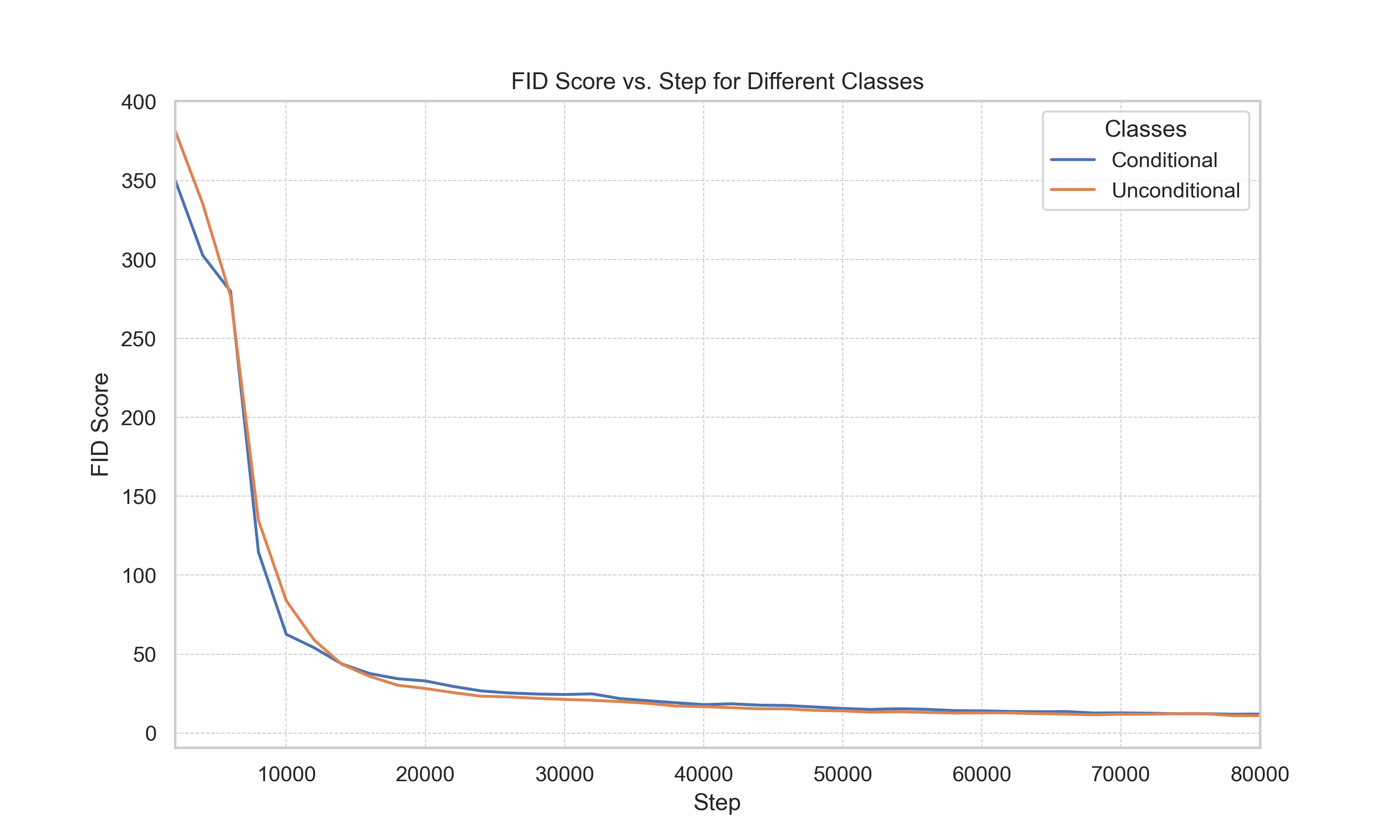}
        \textbf{(a)} Training StyleGAN2 on AFHQ conditionally and unconditionally
    \end{minipage}
    \begin{minipage}{0.48\textwidth}
        \label{fid_number_of_classes_imagnet}
        \centering
        \includegraphics[width=\linewidth]{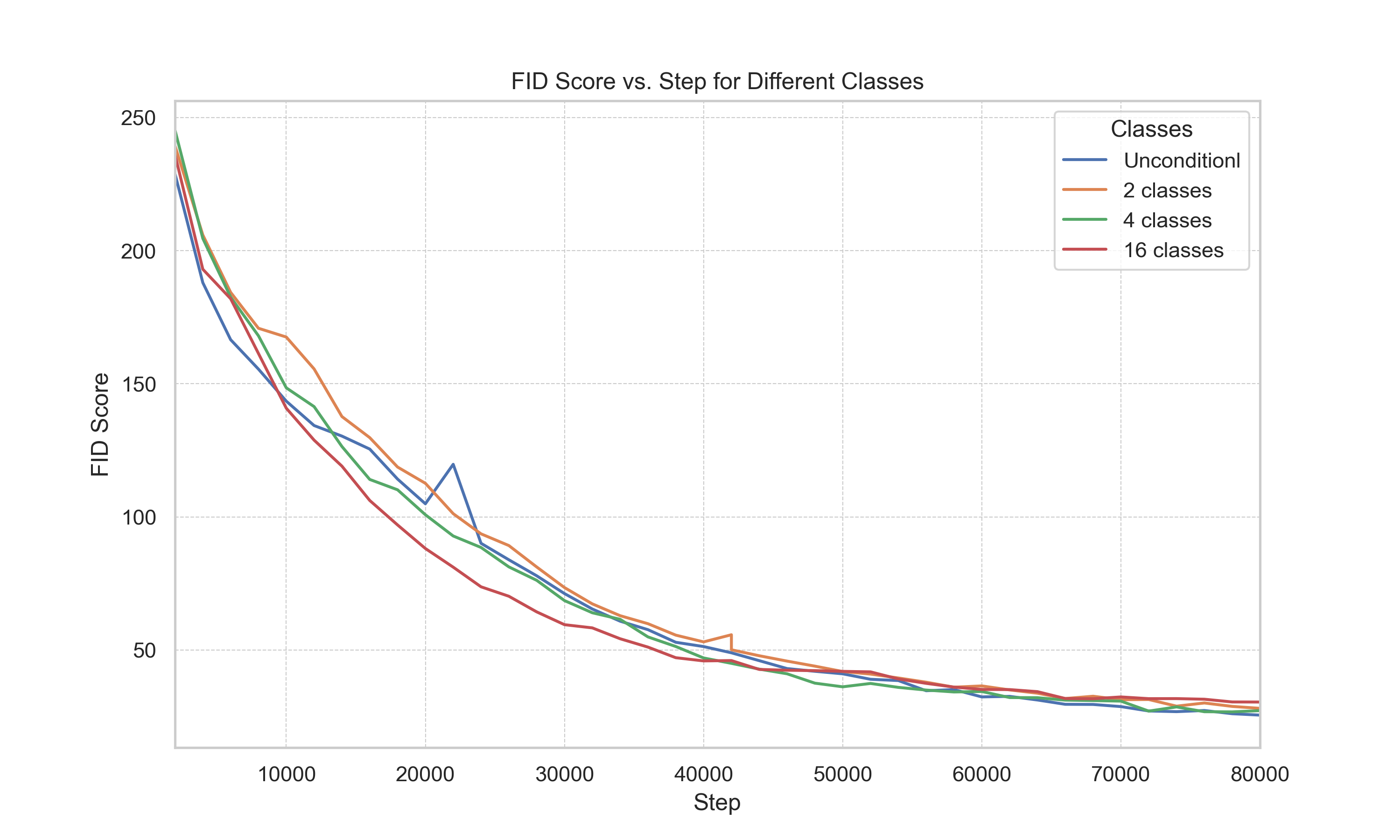}
        \textbf{(b)} Training StyleGAN2 on a subset of Imagnet with different number of classes
    \end{minipage}
    \caption{Training StyleGAN2 on AFHQ and a subset of ImageNet with and without conditioning}
    \label{fig:main}
\end{figure}

In this section, we conduct a comparative analysis of various conditioning techniques across multiple datasets. To ensure fairness and consistency, our evaluation of different architectures leverages StudioGAN~\cite{studiogan}, a PyTorch-based library that provides comprehensive implementations of numerous GAN architectures and conditioning strategies.
We focus on evaluating the proposed architectures on CIFAR-10~\cite{cifar} and a subset of ImageNet~\cite{image_net}. This evaluation maintains the architectures as originally described in their respective works, allowing us to assess their performance under consistent conditions using various metrics.

\subsection{GANs evaluation metrics}
To evaluate the performance of generative models, various metrics were proposed.

\noindent \textbf{Inception Score (IS):} The Inception Score (IS)~\cite{improved_gans,inception_score} is a metric for evaluating the quality of images generated by GANs, leveraging an Inception model~\cite{inception_model} pre-trained on the ImageNet dataset~\cite{image_net}. It quantifies the performance of GANs based on two criteria: the diversity of the generated images across different classes and the confidence of each image's classification. The score is computed by using the Inception model to predict the class distribution for each generated image, assessing both the individual image clarity through the sharpness of its predicted class distribution and the overall diversity by comparing these distributions across all images.

\begin{equation}
\text{IS}(X_t) = \exp\left(\frac{1}{M} \sum_{i=1}^{M} \text{D}_{\text{KL}} \left( p(y|x_i) \| \hat{p}(y) \right) \right)
\end{equation}

Where \( X_t = \{x_1, \dots, x_M\} \) is the image samples we target to evaluate.
\newline
\noindent \textbf{Fr\'{e}chet Inception Distance (FID):} FID \cite{fid} is a metric used to evaluate the quality of images generated by GANs. The FID score calculates the distance between the feature vectors of real and generated images, extracted using an Inception model pre-trained on the ImageNet dataset. Specifically, it computes the Fr\'{e}chet(also known as the Wasserstein-2 distance) between the multivariate Gaussian distributions of the feature vectors of the real and generated images.

\begin{equation}
\text{FID}(X_s, X_t) = \|\mu_s - \mu_t\|_2^2 + \text{Tr} \left(\Sigma_s + \Sigma_t - 2(\Sigma_s \Sigma_t)^{\frac{1}{2}} \right)
\end{equation}

where \( \mu \) and \( \Sigma \) are the mean vector and covariance matrix of the features, and the subscripts \( s \) and \( t \) denote the source and target, respectively.
\newline
\noindent \textbf{Precision and Recall (PR):} 
Precision and recall metrics serve to evaluate the quality and variety of images produced by generative models, relying on comparisons between the distributions of real and generated images. Precision measures the degree to which the generated images resemble the real data distribution, indicating the accuracy of the images produced. In contrast, recall assesses how well the range of real images is represented within the generated images' distribution, reflecting the model's ability to capture the diversity of the real dataset. Precision and recall are defined as follows:

\begin{equation}
\text{Precision} := \frac{1}{M} \sum_{j=1}^{M} \mathbf{1}_{Y_j \in \text{Manifold}(X_s)}
\end{equation}
\begin{equation}
\text{Recall} := \frac{1}{N} \sum_{i=1}^{N} \mathbf{1}_{X_i \in \text{Manifold}(X_t)}
\end{equation}

Where $N$ and $M$ are the number of real and fake samples, the manifolds are defined as:
\begin{equation}
\text{Manifold}(X_1, \dots, X_N) := \bigcup_{i=1}^{N} B(X_i, \text{NND}_k(X_i))
\end{equation}

where \( B(x, r) \) is the sphere in \(\mathbb{R}^D\) around \( x \) with radius \( r \).
\(\text{NND}_k(X_i)\) denotes the distance from \( X_i \) to the \( k^{\text{th}} \) nearest neighbour among \( \{X_i\} \), excluding itself.

\noindent \textbf{Density and Coverage (PR):} 

In \cite{coverage_density}, it was shown that the process of constructing manifolds using the nearest neighbor function is sensitive to outlier samples, which frequently leads to an overestimated representation of the distribution. To address this overestimation issue, they introduced the Density and Coverage metrics, which correct the problem by incorporating sample counting. These metrics are mathematically defined as follows:
\begin{equation}
\text{Density} := \frac{1}{kM} \sum_{j=1}^{M} \sum_{i=1}^{N} \mathbf{1}_{Y_j \in B(X_i, \text{NND}_k(X_i))}
\end{equation}

\begin{equation}
\text{Coverage} := \frac{1}{N} \sum_{i=1}^{N} \mathbf{1}_{\exists j \text{ s.t. } Y_j \in B(X_i, \text{NND}_k(X_i))}
\end{equation}

Where k is for the k-nearest neighbourhoods.

\begin{figure}[t]
    \label{fig:fid_vs_time}
    \centering
    \begin{minipage}{0.48\textwidth}
        \centering
        \includegraphics[width=\linewidth]{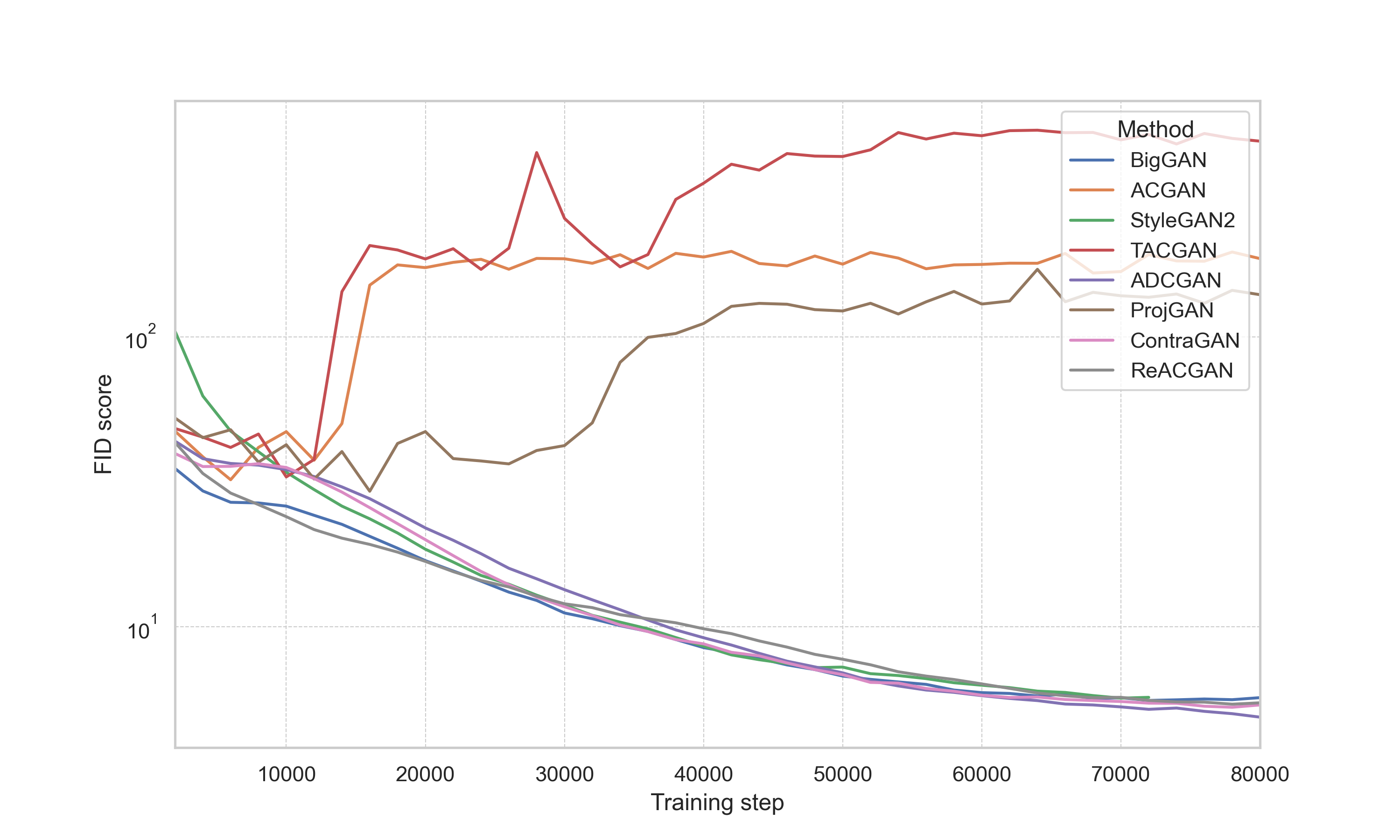}
        \textbf{(a)} CIFAR 10
    \end{minipage}
    \begin{minipage}{0.48\textwidth}
        \label{fig:fid_carnivors}
        \centering
        \includegraphics[width=\linewidth]{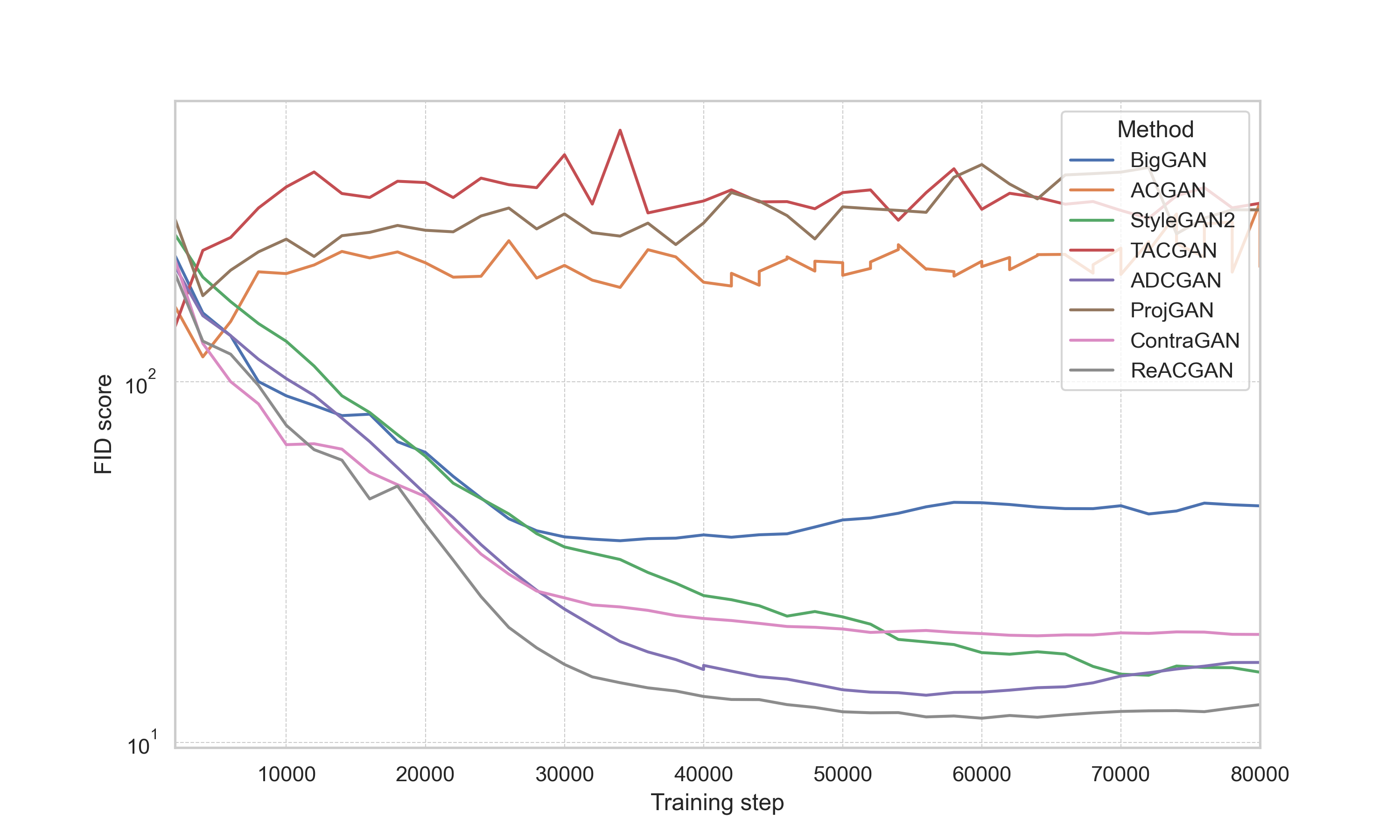}
        \textbf{(b)} Carnivores dataset 
    \end{minipage}
    \caption{FID scores VS training setps for different cGANs architectures trained on CIFAR 10 and Carnivores }
    \label{fig:main}
\end{figure}

\subsection{Evaluation on CIFAR 10 and Carnivores dataset}

First, we evaluated the performance of eight representative cGAN architectures on the \textbf{CIFAR-10} dataset. Table~\ref{tab:cifar_10} presents the scores achieved across three separate runs for each architecture, reported at 80,000 training steps.

In table~\ref{tab:cifar_10}, we observe that \textbf{ACGAN, BigGAN, and TAC-GAN} exhibit nearly identical performance. Additionally, these architectures show a tendency to collapse after approximately the first 10,000 training steps, as illustrated in Fig.~\ref{fig:fid_vs_time}. For the other methods, their performance is relatively similar on this specific dataset, despite belonging to different conditioning families (ADC-GAN, ContraGAN, and StyleGAN2).

Subsequently, in table~\ref{tab:cifar_10}, we evaluated these architectures on a subset of the ImageNet dataset (Carnivores), which consists of larger images sized 128 × 128 pixels. We observe that \textbf{ACGAN, TAC-GAN, and ProjGAN} struggle to accurately learn the distribution of the Carnivores dataset, as indicated by their high FID values. Furthermore, these models tend to collapse earlier than they do on CIFAR-10, highlighting the challenges they face in scaling to larger images.

In contrast, the other architectures demonstrated strong performance without collapsing. BigGAN achieved a higher FID score, while ReACGAN slightly outperformed the other models. Furthermore, as illustrated in fig~\ref{fig:fid_vs_time}, these architectures exhibit greater robustness against mode collapse.

\subsection{Conditional vs unconditional image generation}

Can we achieve better image generation by providing the network with the class of each image? Intuitively, conditioning the model on class labels can be viewed as providing it with additional information, which has the potential to enhance the quality of the generated images. To investigate this, we trained two StyleGAN2 models using the AFHQ~\cite{stylegan2} dataset, which comprises three distinct classes. The first model was trained without conditioning (unconditionally), while the second model was trained with conditioning, where class labels were provided. The FID scores for both models are presented in Fig.~\ref{fid_number_of_classes_afhq}. The results indicate that the two curves are almost similar, suggesting that in this particular case, conditioning does not significantly impact the quality of the generated images.

Given that the AFHQ dataset contains only three distinct classes, we created a specialized dataset derived from ImageNet to gain a deeper understanding of how varying the number of classes affects model performance. In this dataset, the total number of images (resized to 128x128) was kept constant, while the number of classes was varied. Fig.~\ref{fid_number_of_classes_imagnet} illustrates the FID score curves for each training scenario. We began with unconditional training and incrementally increased the number of classes (2, 4, 8, 16). Throughout these experiments, the same set of images was used in all training sessions, ensuring that the model consistently learned from the same distribution.

Our observations reveal two key insights: first, all training runs eventually converge to a similar FID score by the end of the training. Second, we observe that as the number of classes increases, the convergence rate accelerates. This suggests that conditioning can speed up the convergence process in GANs, especially when dealing with datasets containing a large number of classes.

\begin{figure}[h]
    \centering
    \includegraphics[width=0.8\linewidth]{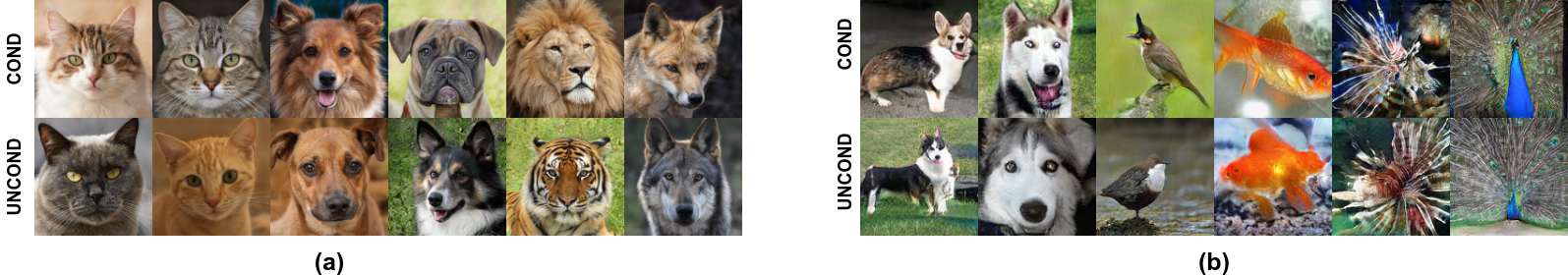}
    \caption{Generated samples from StyleGAN2, both conditionally and unconditionally. On the left, the model was trained on AFHQ (3 classes), while on the right, it was trained on a subset of ImageNet (16 classes). We can notice that the quality of the generated images is almost the same in both settings.}
    \label{fig:cond_uncond}
\end{figure}

\section{Conclusion}

In this survey, we have explored various methods for conditioning Generative Adversarial Networks (GANs), focusing on three primary families of techniques: Auxiliary-classifier based, Projection based, and Contrastive learning based. Each of these approaches offers unique mechanisms to enhance the control and quality of generated images, addressing different challenges inherent in GAN training.

Auxiliary-classifier based methods, such as AC-GAN and its variants, integrate additional classifiers to improve class-specific image generation. These methods have demonstrated improvements in image quality but often struggle with issues like mode collapse and reduced diversity as the number of classes increases. Subsequent enhancements, like the Twin Auxiliary Classifier GAN (TAC-GAN), have sought to mitigate these issues by refining loss functions and incorporating mutual information estimators.

Projection-based discriminators offer a novel approach by conditioning on the inner product between embedded conditional vectors and feature vectors. This family of methods enhances training stability and performance without requiring additional classifiers. Techniques in this category have proven effective in maintaining inter-class diversity and generating high-fidelity images.

Contrastive learning based methods, exemplified by models like ContraGAN and ReACGAN, address the limitations of earlier approaches by focusing on data-to-data relations. These techniques use contrastive losses to maintain diversity and mitigate mode collapse.

Through a comparative analysis of datasets, including CIFAR-10 and a subset of the ImageNet dataset (Carnivores), we found that enhancements to GANs significantly benefit conditioning methods. Notably, contrastive learning-based architectures, projection-based techniques, and auxiliary classifier methods consistently achieve low FID scores.

In conclusion, the advancements in GAN conditioning techniques have significantly enriched the capabilities of generative models. The insights gained from this body of work are invaluable for guiding future research and applications in generative modeling. By continuing to innovate and refine these methods, we can unlock new potentials in GANs, paving the way for groundbreaking applications and more controlled, high-quality image generation.

\bibliography{main}
\bibliographystyle{tmlr}

\end{document}

%% file: math_commands.tex
\usepackage{amsmath,amsfonts,bm}

\def\eqref#1{equation~\ref{#1}}

\def\1{\bm{1}}

\DeclareMathAlphabet{\mathsfit}{\encodingdefault}{\sfdefault}{m}{sl}
\SetMathAlphabet{\mathsfit}{bold}{\encodingdefault}{\sfdefault}{bx}{n}

